\documentclass[letterpaper, 10 pt, conference]{ieeeconf}  

\IEEEoverridecommandlockouts                              

\usepackage{blindtext}

\usepackage{graphics} 
\usepackage{booktabs}
\usepackage{cite}
\usepackage{balance}

\usepackage{graphicx} 
\usepackage{verbatim}
\usepackage{xcolor}
\usepackage{subcaption}

\usepackage[keeplastbox]{flushend}

\usepackage{hyperref}

\usepackage{multirow}

\usepackage{array}

\usepackage{color}
\usepackage{colortbl}
\definecolor{todo-red}{RGB}{200,12,12}
\definecolor{green4}{RGB}{0,128,0}

\newcommand{\reffig}[1]{Fig.~\ref{#1}}

\newcommand{\reftab}[1]{Table~\ref{#1}}
\newcommand{\refsec}[1]{Section~\ref{#1}}

\usepackage{amsmath}
\DeclareMathOperator*{\argmax}{arg\,max}

\title{\LARGE \bf
LandmarkBoost: Efficient Visual Context Classifiers \\ for Robust Localization
}

\author{\authorblockN{
Marcin Dymczyk\authorrefmark{1}, 
Igor Gilitschenski\authorrefmark{3},
Juan Nieto\authorrefmark{1}, 
Simon Lynen\authorrefmark{1}\authorrefmark{2}, 
Bernhard Zeisl\authorrefmark{2}, 
and Roland Siegwart\authorrefmark{1}
}
\authorblockA{\authorrefmark{1}Autonomous Systems Lab, ETH Z\"urich, 
\authorrefmark{2}Google Inc., Z\"urich,
\authorrefmark{3}CSAIL, MIT
}
}

\begin{document}

\maketitle
\thispagestyle{empty}
\pagestyle{empty}

\begin{abstract}
The growing popularity of autonomous systems creates a need for reliable and efficient metric pose retrieval algorithms.
Currently used approaches tend to rely on nearest neighbor search of binary descriptors to perform the 2D-3D matching and guarantee realtime capabilities on mobile platforms.
These methods struggle, however, with the growing size of the map, changes in viewpoint or appearance, and visual aliasing present in the environment.
The rigidly defined descriptor patterns only capture a limited neighborhood of the keypoint and completely ignore the overall visual context.

We propose LandmarkBoost -- an approach that, in contrast to the conventional 2D-3D matching methods, casts the search problem as a landmark classification task.
We use a boosted classifier to classify landmark observations and directly obtain correspondences as classifier scores.
We also introduce a formulation of visual context that is flexible, efficient to compute, and can capture relationships in the entire image plane.
The original binary descriptors are augmented with contextual information and informative features are selected by the boosting framework.
Through detailed experiments, we evaluate the retrieval quality and performance of LandmarkBoost, demonstrating that it outperforms common state-of-the-art descriptor matching methods.
\end{abstract}


\section{Introduction}


Visual localization has become one of the core functionalities in robotics, being widely deployed on mobile platforms and various mobile devices.
High-frequency metric 6-DoF pose estimates expressed in a global map frame enable or facilitate multiple tasks such as navigation, path-planning, obstacle avoidance, or multiagent collaboration.
Frameworks that are accurate over long periods of time are also key for lifelong visual teach and repeat.
These applications require reliable visual localization in the presence of illumination changes, varying weather conditions, or self-similarity of the environment.


Existing localization systems used in robotics usually rely on 2D-3D matching for precise metric pose estimation.
This correspondence search is based on handcrafted features, often binary BRISK~\cite{leutenegger2011brisk} or FREAK~\cite{alahi2012freak} descriptors, to enable realtime queries, even on platforms with limited computational resources.
Their matching performance, however, suffers with the growing size of the database, visual aliasing~\cite{cummins2008fab}, or changes in illumination and viewpoint.
Intuitively, the small image patches covered by descriptor patterns are not expressive when reaching the city or even the building scale.
The high dimensionality makes the search particularly challenging when covering large environments, and requires either approximate methods~\cite{norouzi2012fast} or projection into different spaces~\cite{lynen2014placeless}.
%



More recent algorithms attempt to improve the performance of binary descriptors by checking for geometric consistency or by augmenting them with additional information, such as a broader neighborhood or semantic labels.
Another approach is to use larger patches or the entire frame and learn the features with end-to-end methods.
This leads, however, to much higher computational demands, basically requiring a costly, heavy and power-demanding GPU to be installed onboard.
Additionally, most state-of-the-art systems merely consider place recognition, i.e. retrieving a topological location such as a map image, but do not provide an exact 6-DoF pose.
This does not fulfill the requirements of robotic applications that rely on metric output.


 \begin{figure}[t]
     \centering
     \caption*{
     \begin{tabular}{>{\centering}m{0.45\columnwidth} >{\centering}m{0.45\columnwidth}}
     Query frame & Closest map frame
     \end{tabular}}
     \vspace{-0.1cm}
     \includegraphics[width=1.0\columnwidth]{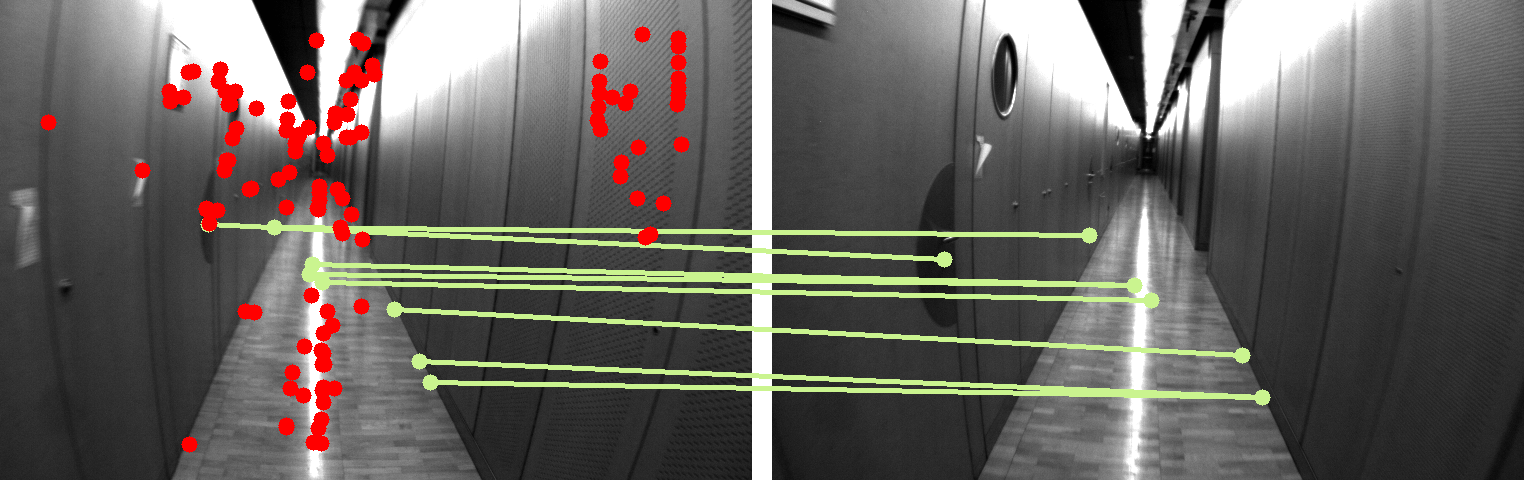}
     \vspace{-0.5cm}
     \caption*{BRISK}
     \includegraphics[width=1.0\columnwidth]{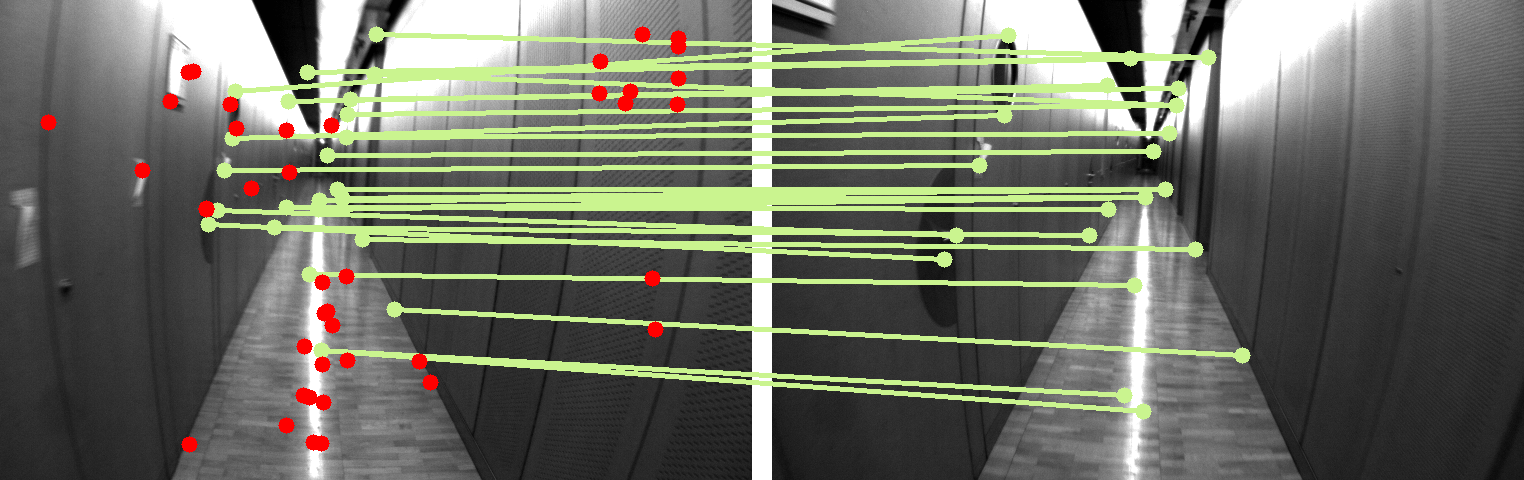}
     \vspace{-0.5cm}
     \caption*{LandmarkBoost}
     \caption{Matching result for BRISK descriptors and the proposed LandmarkBoost framework in a very self-similar environment.
     The left frame is a query image, the right one is the closest map frame.
     Red points in the query frame denote outliers, i.e. returned matches that failed geometric verification.
     Thanks to the inclusion of context information LandmarkBoost retrieves more geometrically consistent matches and fewer outliers.}
     \label{fig:teaser_image}
     \vspace{-0.6cm}
 \end{figure}

We propose LandmarkBoost, a system that addresses the specific needs of mobile robotics for large-scale metric visual-based localization while keeping a limited computational cost of retrieval.
Our method augments binary descriptors with visual context to build a comprehensive descriptor of the observations and improve the 2D-3D matching performance.
The augmented descriptors are passed to a boosted classifier, effectively casting the matching problem as a classification.
The algorithm returns 2D-3D correspondences that can be directly consumed by a 6-DoF pose estimation engine.
The new features and a retrieval engine better handle the typical failure cases, such as visual aliasing, and improve the matching results, as in \reffig{fig:teaser_image}.
The contributions of this work can be summarized as follows:
\begin{itemize}
    \item We propose a visual context formulation that augments the binary descriptors and permits to improve the raw descriptor matching.
    The context is extracted from randomly generated regions that capture co-occurring statistics between landmarks within the entire image plane.
    \item We introduce a boosted classifier that serves as an efficient observation-to-landmark matching engine, but also as a tool to discover useful context relationships between landmarks.
    The optimized implementation of the classifier enables support for large scale maps.
    \item We demonstrate that the runtime of the proposed approach is comparable to the conventional search methods for binary and projected descriptors.
    This confirms our boosted classifier can be used to perform online localization.
    \item We provide an extensive experimental validation that demonstrates our method outperforms the commonly-used binary descriptors both at 2D-3D correspondence search and the resulting pose estimation.

\end{itemize}

\section{Related work}

The principle of 2D-3D matching~\cite{sattler2011fast} followed by a n-point-pose (PnP) solver with RANSAC is a commonly used toolset for metric localization in robotics and mobile devices.
The efficiency and scalability of such approaches~\cite{middelberg2014scalable} requires limited computational resources and permits to estimate a 6-DoF pose with high accuracy.
Further optimizations to the pipeline~\cite{lynen2015get} let us use such methods in real-time, making them particularly suitable for mobile robotics, enabling new applications such as collaboration between agents~\cite{fankhauser2016collaborative}.

The quality of retrieved 3D landmark matches for a query 2D keypoint directly affects the precision and recall of pose estimation.
Below, we present selected approaches that are used to improve the raw descriptor matching output.

\textbf{Voting and filtering:} The raw correspondence search result can be improved by filtering matches or reranking the list of candidates using some additional information.
In~\cite{jegou2008hamming}, {\em J\'egou et al.} introduce a Weak Geometric Consistency test while building a list of matches that checks if scale and angle of the retrieved matches are consistent with the query.
A related approach is presented by {\em Zeisl et al.}~\cite{zeisl2015camera}, where spatial verification is formulated as a Hough voting problem that works in linear time.
An alternative approach is proposed by {\em Lynen et al.}~\cite{lynen2014placeless}, where the raw descriptor matches are used to construct a voting space to only return the matches from high vote density regions.
This method was further improved by a probabilistic approach to voting  in~\cite{gehrig2017visual}.
All these approaches are complementary to LandmarkBoost and can be used to further refine the candidate list returned by our algorithm.

\textbf{Image retrieval:} Another group of approaches considers entire images instead of local features.
Some of these methods depend on aggregating descriptors of local features, e.g. bag-of-words~\cite{csurka2004visual} or VLAD~\cite{jegou2010aggregating}.
The image retrieval approaches can then be used to model places~\cite{cummins2008fab} and avoid potentially misleading features.
They can also be used to obtain a localization prior to constrain the search space for 2D-3D correspondences~\cite{irschara2009structure}.
Another family of methods describe the entire image without relying on keypoints, using handcrafted or learned methods, examples including GIST~\cite{oliva2001modeling} and DIR~\cite{gordo2016deep} respectively.
These algorithms, however, rely on relatively expensive computations and large descriptor dimensionalities.
An interesting approach is presented in~\cite{weyand2016planet}, where a classifier is used to assign an image to one of over 26k geographic locations worldwide.
It proves viability of classification methods for place recognition.

\textbf{Temporal sequences:} The localization queries can also be extended in the temporal dimension.
Intuitively, using sequences of frames helps to distinguish places and avoid mistakes, e.g. caused by visual aliasing.
Sequence-based methods can rely on traditional image comparison measures~\cite{milford2012seqslam} or using long short-term memory (LSTM) networks~\cite{clark2017vidloc}.
Using ordered sets of frames, however, raises a question about robustness to trajectory changes and viewpoint variations, as small shifts of field-of-view can disrupt the performance~\cite{sunderhauf2013we}.

\textbf{Landmark covisibility:} Yet another group of approaches tries to benefit from analyzing co-occurrence statistics of groups of landmarks.
The assumption is that while it might be easy to confuse a single landmark observation, it is much less probable when dealing with a covisibility subgraph.
{\em Mei et al.}~\cite{mei2010closing} and later {\em Stumm et al.}~\cite{stumm2016building, stumm2016robust} present approaches that rely on landmark covisibility statistics to find reappearing patterns in the graph space.
Similarly, in~\cite{lowry2018logos} orientation and scale of neighboring features are used.

\textbf{Visual context and classifiers:} Instead of landmark covisibility, we can focus just on a neighborhood of a point feature in the image plane, called visual context.
In~\cite{loquercio2017efficient}, {\em Loquercio et al.} suggest to combine point features with a local neighborhood of a fixed size.
{\em Zhang et al.}~\cite{zhang2011image} propose to build geometry-preserving visual phrases, that capture visual word co-occurrences and most importantly their spatial layout even including distant relationships in the image plane.
It can be considered an extension to~\cite{yuan2007discovery} where just local patterns were discovered.
A depart from point matching is suggested in~\cite{mcmanus2014scene}\cite{linegar2016made}, where a bank of SVMs, operating on mid-level features, is used to obtain associations to the prior map.
Unfortunately, while improving robustness, this approach reduces the metric accuracy of the pose estimates.

The method proposed in this paper aims to combine the advantages of some of the aforementioned approaches.
We introduce a notion of visual context that is able to represent even long-range co-occurring feature patterns, implicitly capturing landmark covisibility instead of modelling it~\cite{stumm2016building}.
The context is described by sampling random region candidates and deciding during the training which of them contain useful information.
This approach resembles the banks of SVMs in~\cite{mcmanus2014scene}, but depends on region embeddings and simple decision stumps.
Finally, we use a multi-class classifier to directly obtain 2D-3D matches which yields a precise metric localization.

\section{Methodology}

We present LandmarkBoost -- a localization algorithm that uses a visual context and boosted classifiers to obtain 2D-3D matches between query keyframes and prior map landmarks.
We assume the prior map to be a structure-from-motion (SfM) model that contains covisibility information, detected keypoints, extracted features and 3D landmark positions.
Additionally, we augment the SfM model with a concept of visual context that takes into account reoccurring feature patterns to improve the matching.
In \refsec{sec:context} we present our approach to model the visual context.
\refsec{sec:jboost} introduces the learning framework based on Jointboosting~\cite{torralba2004sharing}, with certain specific adaptations to the task of 2D-3D matching.
The boosted classifiers are area-specific and need to be pretrained offline, using both the conventional features and visual context embeddings as input.
The framework supports a large number of 3D landmarks and lets us perform visual context feature selection.
Finally, in \refsec{sec:classification}, we present an efficient approach to perform the classification online.

\subsection{Modeling visual context}
\label{sec:context}

\begin{figure}
    \centering
    \includegraphics[width=\columnwidth]{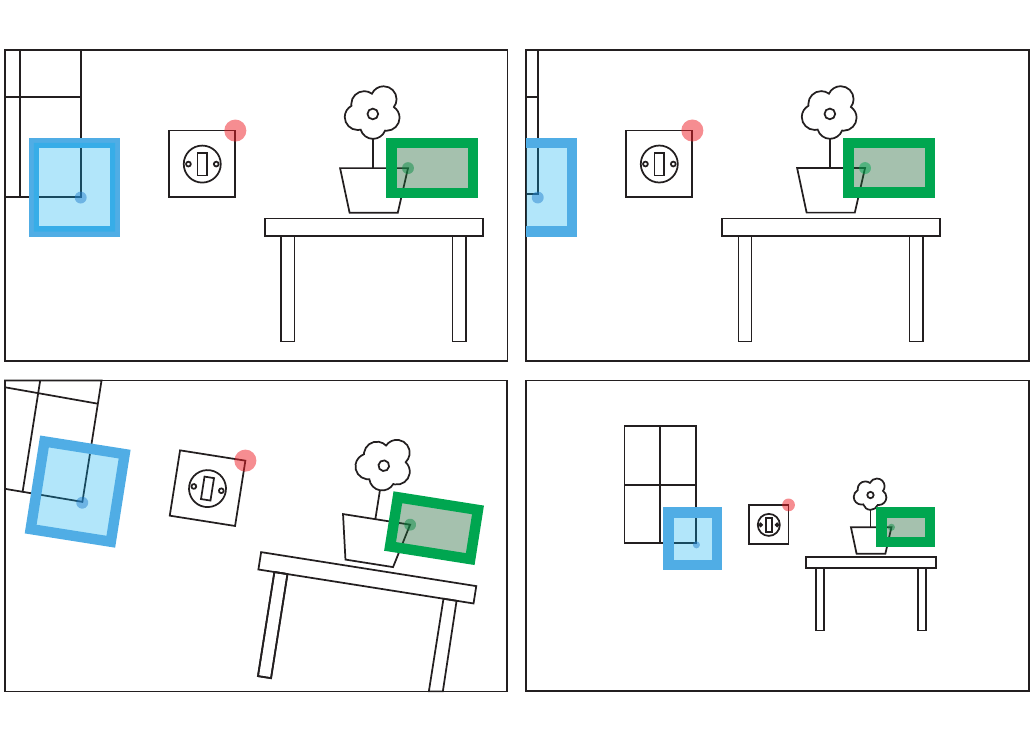}
    \caption{ 
    Context regions are anchored to a keypoint (red point) so that they move with the keypoint (top right). 
    They model even distant relationships as they can be constructed anywhere in the image plane.
    They are rotated according to the gravity direction (bottom left).
    Finally, they are scaled using the keypoint detector scale to guarantee scale invariance (bottom right).
    }
    \label{fig:region_invariance}
    \vspace{-4mm}
\end{figure}

We propose to describe the visual context by means of context regions -- areas in the image plane anchored to the point features.
Those regions can be used to extract context embeddings that will be used to build feature vectors for context learning.
The regions are common for all landmarks, i.e. the same set is applied to all observations.
Initially, a large number of randomized regions is generated to guarantee a good coverage of the possible visual context locations.
Then, we mine for the informative ones during subsequent training.

\subsubsection{Context regions}

Typically, feature-point descriptors are only extracted from the direct neighborhood of the keypoint location.
For BRISK descriptors, the pattern covers a radius of 16~pixels.
That is, however, often not sufficient to guarantee reliable matching as many corners look alike.
Visual aliasing and identical objects appearing in many places within one environment result in false matches and frequent localization failures.

We propose the use of context regions anchored to the keypoint locations.
Such regions can span much larger areas than the typical descriptor patterns.
Their locations in the image are \textbf{relative to the keypoint location} and expressed in the normalized undistorted image plane to factor out the influence of lens effects.
Additionally, the regions are scaled according to the keypoint scale to guarantee \textbf{scale invariance} and they are rotated using the gravity direction projected into the frame to ensure \textbf{rotation invariance}.
The invariance properties of the regions are illustrated in~\reffig{fig:region_invariance}.

\subsubsection{Region generation}

As we do not know \emph{a priori} which parts of the keypoint neighborhood are useful for learning the context, our approach aims to generate a large number of regions and only later decide which are actually useful for each landmark.
We use rectangular regions within the image plane, with varying areas, aspect ratios and locations.

While the area of regions is random, we guarantee that each point in the image has a uniform probability of being covered by a region of any admissible area.
In other words, we want to guarantee the region area distribution to be:
\begin{equation}
    Pr({\rm area}) \propto \frac{1}{\rm area}
\end{equation}
Using an inverse transform sampling method, we can sample from uniform distribution $U(m, n)$ and generate a desired region area using the inverse distribution function:
\begin{equation}
    {\rm area} = F^{-1}(u) = \exp(u)
\end{equation}

\subsubsection{Region descriptors}

Regions, when applied to the image, denote an area that contains visual context information.
This information can be extracted as a feature vector and used for context learning.
Our method describes regions using bag-of-words (BoW) embeddings~\cite{csurka2004visual} that are inexpensive to compute, compact and straightforward to interpret.
The BoW vectors are computed per region, based on keypoint descriptors within the region, and then normalized.
The visual vocabulary for BoW generation is trained using the k-means algorithm on an unrelated dataset using the same descriptor type.

\subsection{Training shared classifiers}
\label{sec:jboost}

The proposed method attempts to train a classifier that takes a query feature vector $v$ as an input (which may contain both, conventional descriptors and context embeddings) and outputs a ranked list of matching landmarks from the database.
Our training algorithm is based on Jointboosting by {\em Torralba et al.}~\cite{torralba2004sharing} which satisfies several of our requirements:
\begin{itemize}
    \item the classifier is based on simple decision stumps, which makes evaluation even for tens of thousands of classes feasible and easy to parallelize,
    \item it relies on jointly training the decision stumps and sharing them among a set of classes so even a few observations are enough to build a reliable classifier for a landmark,
    \item the classifier outputs a ranked list of candidates, which can be further filtered,
    \item boosting algorithms inherently perform feature selection that can be used to mine for reliable visual context features.
\end{itemize}

Using a classifier instead of traditional nearest neighbor queries comes, however, at a cost.
We need to train it for a specific set of landmarks, which means each area needs its own classifier.
If a large environment is to be covered, the correct classifier can be preloaded based on a rough pose guess, e.g. GPS reading as proposed in~\cite{mcmanus2014scene}.
Below, we always assume to deal with a single multi-class classifier.

\subsubsection{Constructing training data}

\begin{figure}
    \centering
    \includegraphics[width=0.65\columnwidth]{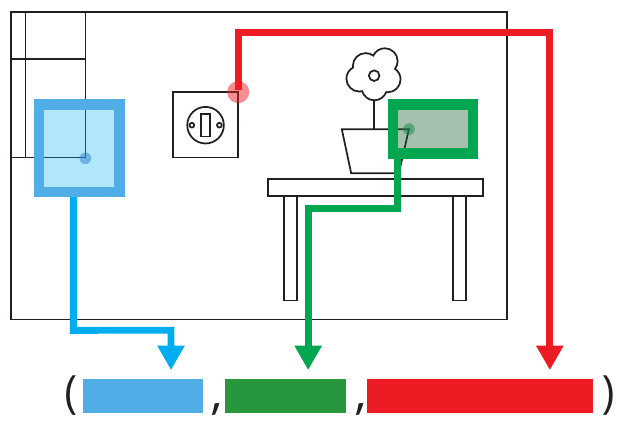}
    \caption{
    The proposed descriptor is a concatenation of descriptors of all context regions (blue and green) and the original keypoint descriptor (red).
    The descriptor of each region is a normalized BoW vector.
    In our experiments, we construct 1,000 regions and use 16 visual words to compute a BoW vector.
    }
    \label{fig:embedding}
\end{figure}

In the boosting framework, each database 3D landmark corresponds to a class.
Each observation of a landmark $c$ corresponds to a single training sample $v_i$ with a label $z_c^i = 1$ and $\forall_{c' \neq c} z_{c'}^i = -1$.
The feature vector for an observation is a concatenation of descriptors of all regions, each of length equal to the BoW vocabulary size.
Additionally, we augment the feature vector with the original keypoint descriptor, see \reffig{fig:embedding}.
By combining the context with the point descriptor, we let the learning algorithm decide which information reduces the misclassification cost more at a given training stage.

\subsubsection{Training procedure}

The boosted classifier $H$, as in~\cite{torralba2004sharing}, is a sum of $M$ weak learners $h_m$:
\begin{equation}
    H(v,c) = \sum_{m=1}^M h_m(v,c)
\end{equation}
where $c$ is the class index and $v$ the feature vector.
At every training round $m$, a new weak learner is added so the $H(v,c)$ is updated to $H'(v,c) = H(v,c) + h_{m}(v,c)$.
The new classifier is expected to reduce the overall weighted misclassification loss.
The original Jointboost formulation follows the gentleboost~\cite{friedman2000additive} algorithm and minimizes the following weighted squared error $J$:
\begin{equation}
    J = \sum_{c \in C} \sum_{i=1}^{N} w_i^c(z_i^c - h_m(v_i,c))^2
\end{equation}
The sample weight $w_i^c$ is specific for a sample $i$ and a class $c$ and evolves during training.
Note that the cost $J$ is a sum over all classes $C$ and all $N$ training samples.
The resulting weak learner function is a binary classifier that operates on a sharing set $S(n)$ and has the form as in~\cite{torralba2004sharing}:
\begin{equation}
    h_m(v, c) = \begin{cases}
    a \cdot \delta(v_i^f > \theta) + b, & \text{if $c \in S(n)$}.\\
    k^c, & \text{if $c \notin S(n)$}.
  \end{cases}
\end{equation}
with parameters $(a, b, k^c, \theta)$ fitted in the boosting round. 
For active classes $c \in S(n)$ the decision stump operates on a dimension $f$ of the feature vector $v_i$ and is a step function at the threshold value $\theta$.
Multiple feature dimensions $f \in F_m$ are tested in a single boosting round.
For inactive classes the value depends on a class-specific constant $k^c$.

\subsubsection{Efficient training of shared classifiers}

The original Jointboost applications included mostly object classification, evaluated on 7- or 21-class datasets.
We want to deploy the algorithm on thousands of classes, one for each 3D landmark, and therefore we introduce a number of optimizations to the original pipeline.

\textit{Negative sample sets per class:} In the original formulation, all samples of class $c$ are used as a positive set  $\mathcal{L}_c^{+}$ and all samples of classes other than $c$ are used as negative samples for a class $c$.
Instead, we introduce a negative $\mathcal{L}_c^{-}$ sample set per class, resulting in the following cost function:
\begin{align}
\begin{split}
    J = \sum_{c \in C} ( & \sum_{\mathcal{L}_c^{+}} w_i^c(1 - h_m(v_i,c))^2  \\  
    + & \sum_{\mathcal{L}_c^{-}} w_i^c(1 + h_m(v_i,c))^2 )
\end{split}
\end{align}
The limited size of the negative set addresses the heavy imbalance of the number of positive and negative samples per class compared to the original formulation.
Additionally, this change significantly reduces the complexity of training, as we only evaluate $|\mathcal{L}_c^{+}| + |\mathcal{L}_c^{-}|$ samples per class.

\textit{Boot-strapping negative samples:} 
After introducing a predefined negative sample set per class $\mathcal{L}_c^{-}$, care needs to be taken to guarantee this set includes hard negative samples.
We use a landmark quantizer to find them, assuming they quantize to the same visual words as the positive samples of the class $c$.
This approach follows the Nearest Neighbor Negatives concept introduced in~\cite{philbin2010descriptor}.
Moreover, we do not add any negative samples that are observations of landmarks located nearby in the metric space -- these might be duplicates of the same landmark that did not get associated correctly by the SfM pipeline.

\textit{Hard negative mining:} 
We further augment the negative set $\mathcal{L}_c^{-}$ while training.
We add samples of classes that were misclassified as positives of a class $c$. 
In this way, the initial negative sample set is getting extended by samples that refine the classifier~\cite{sung1998example}.

\textit{Efficient sharing set initialization:}
In~\cite{torralba2004sharing}, {\em Torralba et al.} propose a greedy approach with a $\mathcal{O}(|C|^2)$ complexity to select the sharing set $S(n)$.
Given the feature dimension $f$, we believe it is possible to initialize the sharing set based on $v_i^f$ values.
Classes that expose consistent behavior (e.g. majority of values are either zeros or very large) can be grouped together to form a candidate set $S(n)$ that would classify them against the rest.
While initial sharing sets are further refined, fewer iterations of the selection loop are necessary.

\textit{Efficient sharing set update:} 
The parameters $a$, $b$ and $k^c$ are refitted~\cite{torralba2007sharing} at each sharing set estimation step.
We propose to incrementally update them. 
E.g., when adding a class $c'$ to the sharing set, we perform an update of $b_S = b_{num} / b_{den}$:
\begin{equation}
    b_S'(f, \theta) = \frac{b_{num} + b_c(f, \theta) w_c'^+(f, \theta)}{b_{den} + w_c'^+(f, \theta)}
\end{equation}
using precomputed values of $b_c(f, \theta)$ and $w_c'^+(f, \theta)$.
While this change might not seem significant, we have to keep in mind that decision stumps are fitted $|F_m| \cdot |C|^2$ times per boosting round so this optimization has a high impact on the total runtime.

\textit{Background class:} The classifiers in~\cite{torralba2007sharing, shotton2006textonboost} never return a ``no match`` answer.
We introduce a background class that consists of samples that belong to untracked keypoints or landmarks not present in the database.
A match to the background class means that the observation could not be matched to any of the known landmarks and should not be used for pose estimation.
This increases the inlier ratio of the PnP+RANSAC algorithm.

\subsection{Localization through Classification}
\label{sec:classification}

The shared decision stumps $h_m$, after training as described in~\refsec{sec:training}, can be used to establish 2D-3D matches, i.e. matches between the keypoints of the query frame and the 3D landmarks stored in the database (a prior map).
The localization works as follows:
\begin{enumerate}
    \item The context embedding is computed by adapting the regions to the keypoint location and scale. 
          For each of the regions, a region descriptor is computed.
    \item The context embeddings of each keypoint are combined with its descriptor to create a feature vector $v$.
    \item The feature vector $v$ is classified by $M$ shared decision stumps:
          \begin{equation}
              c_{best} = \argmax_{c \in C} \sum_{m=1}^{M} h_m(v, c)
          \end{equation}
    \item The keypoint-landmark pairs are translated into 2D-3D matches to estimate the pose using PnP and RANSAC algorithms~\cite{kneip2014opengv}.
\end{enumerate}

The classification scheme presented above evaluates the classifier for all possible classes $C$.
This might impair the performance when the classifier covers large areas with many landmarks.
We propose to mitigate this issue by using an inverted file of database landmark quantizations.
Given a query point descriptor, we can only test the candidates that quantize to the same visual words $w$, effectively reducing the set $C$ to a subset $C_{w} \subset C$.

\section{Experimental evaluation}
\label{sec:experimental}

The experimental evaluation verifies both the performance of the training procedure, as well as landmark and pose retrieval statistics.
In \refsec{sec:training} we describe the training procedure.
Then, in \refsec{sec:landmark_retrieval} we evaluate the landmark retrieval, showing that the suggested approach returns more inlier matches than the baseline methods and that the correct matches are ranked higher.
In \refsec{sec:pose_retrieval} we provide a comprehensive evaluation of the pose retrieval.
Finally, \refsec{sec:runtime} presents a comparison of the runtime and inlier ratio between LandmarkBoost and baseline methods.

We use the implementation of Jointboosting by {\em Kr\"ahenb\"uhl et al.}~\cite{krahenbuhl2011efficient}, with the improvements described in \refsec{sec:jboost}.
The point descriptors are 384-bit BRISK~\cite{leutenegger2011brisk}, extracted by the maplab~\cite{schneider2018maplab} mapping framework.
The proposed algorithm is compared to baseline methods, namely exact binary descriptor (BFMatcher) search implemented by OpenCV~\cite{opencv_library} and exact projected descriptor search using libnabo~\cite{elseberg2012comparison}.
We directly feed the 2D-3D matches provided by the search methods to a localization engine based on PnP+RANSAC provided by opengv~\cite{kneip2014opengv}.
No match clustering or majority voting is used to avoid distorting the results of raw correspondence search.

\begin{figure}
    \centering
    \includegraphics[width=0.45\columnwidth]{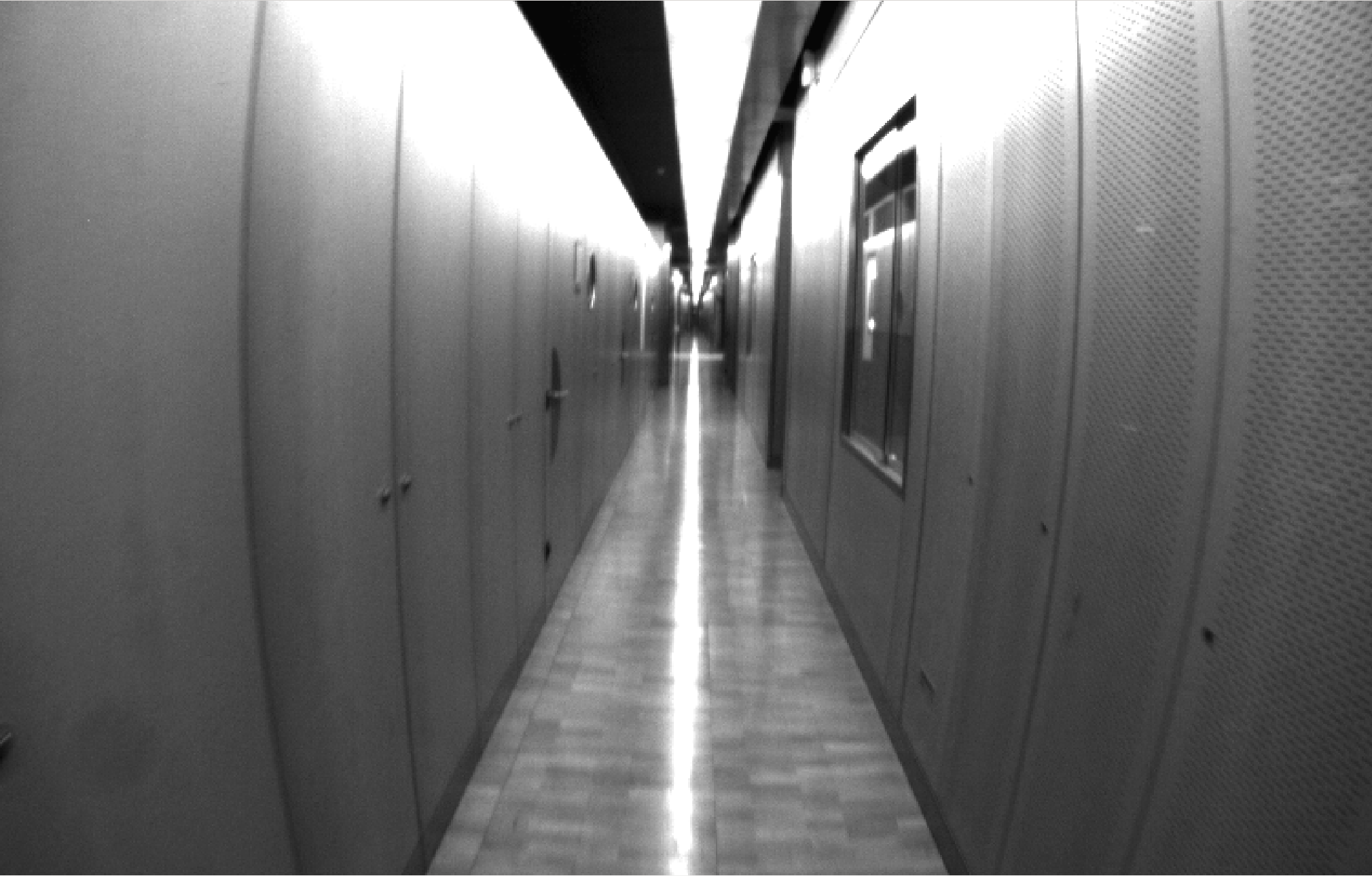}
    \includegraphics[width=0.45\columnwidth]{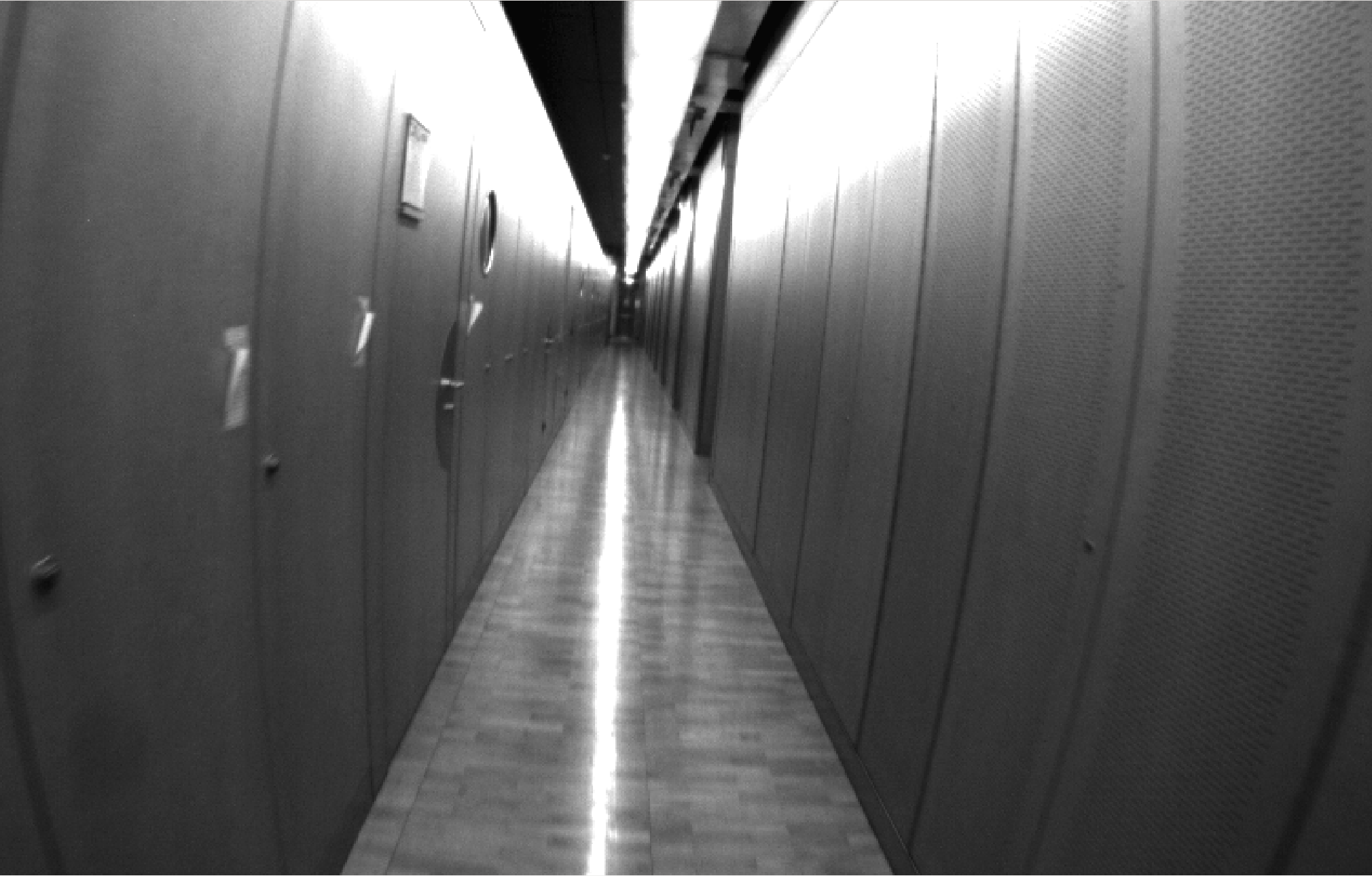}
    \caption{
    A non-matching pair of frames from the CLA F dataset.
    Even though they look alike, the places are more than $20$m apart.
    Using just keypoint descriptors leads to a localization failure due to the extreme visual aliasing.
    Visual context helps to disambiguate the locations -- small cues let us reject false positive landmark matches.
    }
    \label{fig:cla}
\end{figure}

The evaluation is performed on two sets of datasets: very self-similar indoor environment (see \reffig{fig:cla}) and a large-scale outdoor dataset.
The details of the datasets are given in \reftab{table:dataset}.
For each set of datasets, all but one are used to build a database (i.e. build a kd-tree or train a classifier) and the remaining one for evaluation.
The goal is to retrieve relevant descriptors or localize each frame of the evaluation dataset.

\begin{table}
\centering
\begin{tabular}{l>{\hspace{-6pt}} c>{\hspace{-6pt}} c>{\hspace{-6pt}} c>{\hspace{-6pt}} c>{\hspace{-6pt}} c}
				\toprule
				Dataset				& Notes & Distance & Descriptors & Landmarks & Frames \\ \midrule
				\multirow{2}{*}{CLA F}  & Train & 221m & 405k & 41.9k         & 1737 \\
				                              & Eval & 199m & 465k   & 39.5k & 1849 \\ \midrule
				\multirow{2}{*}{Zurich Old Town} & Train & 2441m & 583k   & 35.2k & 2009 \\
				                                 & Eval  &  438m & 243k  & 15.2k  & 565\\ \bottomrule
\end{tabular}
\caption{Train and evaluation datasets.
            The datasets are public and available with the maplab mapping framework~\cite{schneider2018maplab}.
			%
			%
			}
\label{table:dataset}
\end{table}

\subsection{Classifier training}
\label{sec:training}

\begin{figure}
    \centering
    \includegraphics[width=\columnwidth]{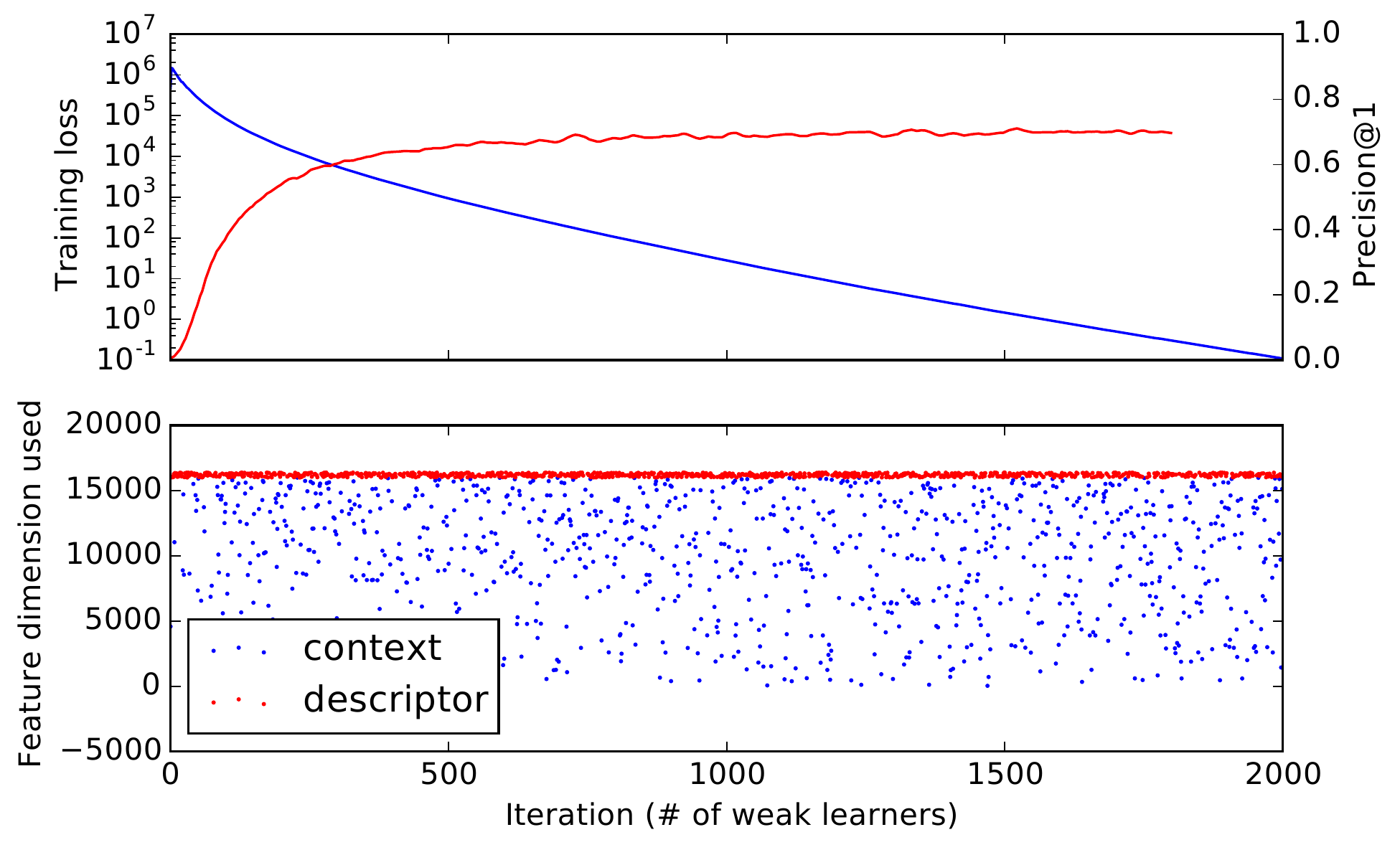}
    \caption{
    The training log of a 7,500-landmark classifier.
    The boosting algorithm successfully reduced the training cost (\textcolor{blue}{blue}) at each iteration.
    The precision@1 (\textcolor{red}{red}) of the evaluation set classification steadily grows to reach the level of 72.8\%.
    The feature vectors contain 16,000 context dimensions (1,000 regions with 16 visual words each) and 384 descriptor dimensions.
    Overall, 985 context features and 1,015 descriptor features were selected by the algorithm.
    Some of the feature dimensions were used by multiple decision stumps and can be combined to speed up the classification.
    }
    \label{fig:training}
\end{figure}

The classifier for the CLA F dataset was trained on 7,500 best landmarks, selected by landmark quality~\cite{dymczyk2015gist, dymczyk2015keep}.
The training fitted 2,000~decision stumps, used 1,000~regions and evaluated 500~features at each round.
Further increasing the region count did not bring significant improvements to the matching performance.
The training progress, that took about 8~hours, is depicted in \reffig{fig:training}.
The fitted weak learners used both keypoint descriptors as well as context embeddings.
This indicates the provided visual context contains useful information that helps to establish 2D-3D matches.

\begin{figure}
\centering
    \includegraphics[width=\columnwidth]{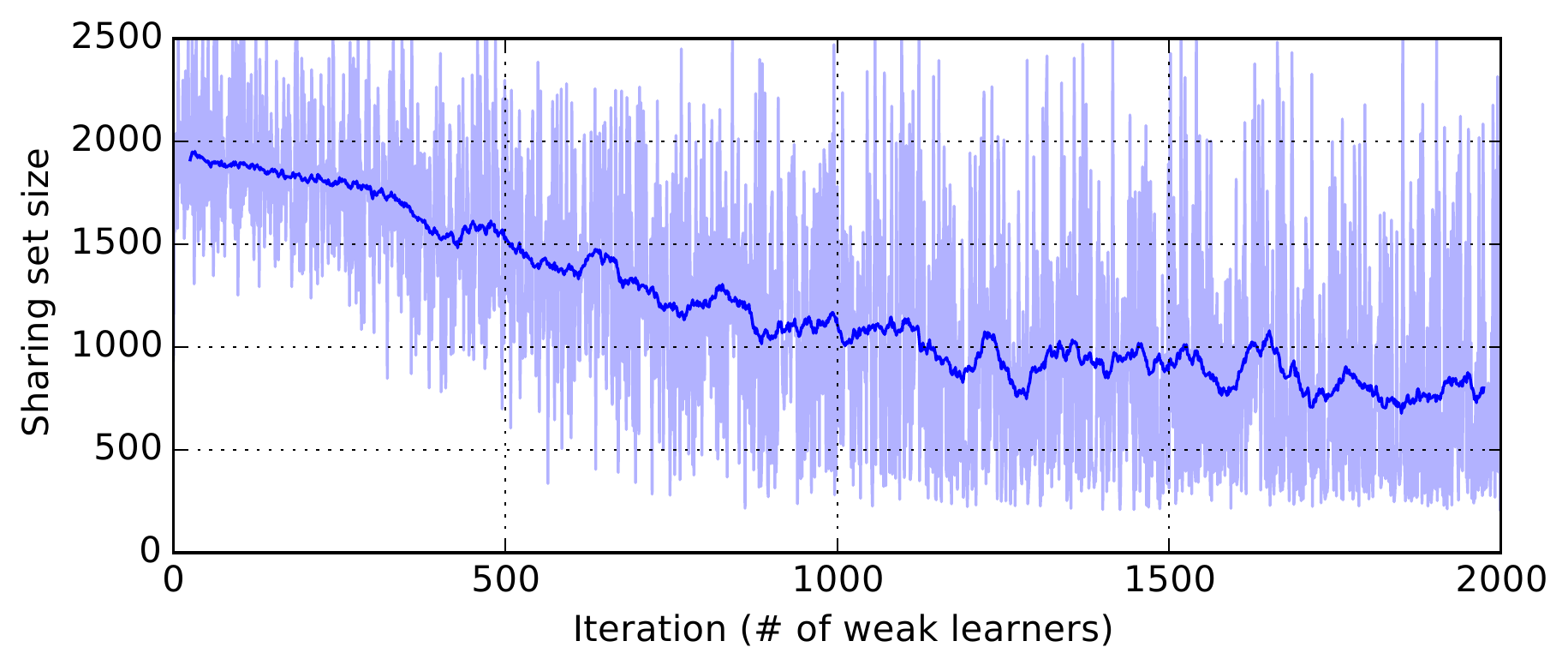}
    \caption{ 
    Evolution of the sharing set size for 5,000 landmarks.
    The size of the sharing set over the boosting rounds follows the intuition.
    Initially, the largest reduction of the misclassification cost is caused by large sharing set, approximately dividing the class set $C$ in two.
    As the training progresses, more fine-grained splits are prevalent.
    The noisy nature of the sharing set size is a result of random candidate feature dimensions $F_m$, greedy sharing set updates and evolving sample weights $w_i^c$.
    }
    \label{fig:sset}
\end{figure}

We have also evaluated the sharing set size over the boosting iterations.
Intuitively, initial sharing sets should contain about 50\% of classes as such splits permit to reduce the classification cost most efficiently.
Later on, the sharing set size is getting smaller as more detailed splits are necessary to improve the accuracy.
This expected behavior indeed takes place when training, as depicted in \reffig{fig:sset}.

All of the above classifiers as well as all the subsequent evaluations use a BoW vocabulary that contains 16 visual words.
This has proven to be an optimal value for descriptor retrieval as ilustrated in \reftab{tab:vw}.

\begin{table}
\centering
\begin{tabular}{l c c c c c c}
				\toprule
						    & 4vw & 8vw & 16vw & 32vw & 64vw & 128vw \\ \midrule
				Precision@1 & 0.339 & 0.419 & \textbf{0.744} & 0.703 & 0.644 & 0.567 \\ \midrule
				MRR         & 0.361 & 0.598 & \textbf{0.851} & 0.825 & 0.786 & 0.7246 \\ \bottomrule
\end{tabular}
\caption{Precision@1 and Mean Reciprocal Rank (MRR) of the Jointboost classifier for selected numbers of visual words (vw).
            The classifier was trained solely on the visual context embeddings, without the raw descriptor part.
            The quality of retrieval is affected by the BoW visual vocabulary size used to describe the regions.
            Too few visual words means low context embedding uniqueness, too many lead to quantization errors.
            }
\label{tab:vw}
\end{table}

\subsection{Landmark retrieval}
\label{sec:landmark_retrieval}

In this section, the quality of the returned 2D-3D matches is evaluated.
The precision and recall of the landmark retrieval has a direct influence on the quality and runtime of subsequent pose retrieval.
A large fraction of false matches (outliers) might prevent the pose estimation algorithms to output a valid solution within a limited time.
We are therefore interested in the precision of the first retrieved candidate, but also the rank of the true positive.

\begin{figure}
    \centering
    \includegraphics[width=1.0\columnwidth]{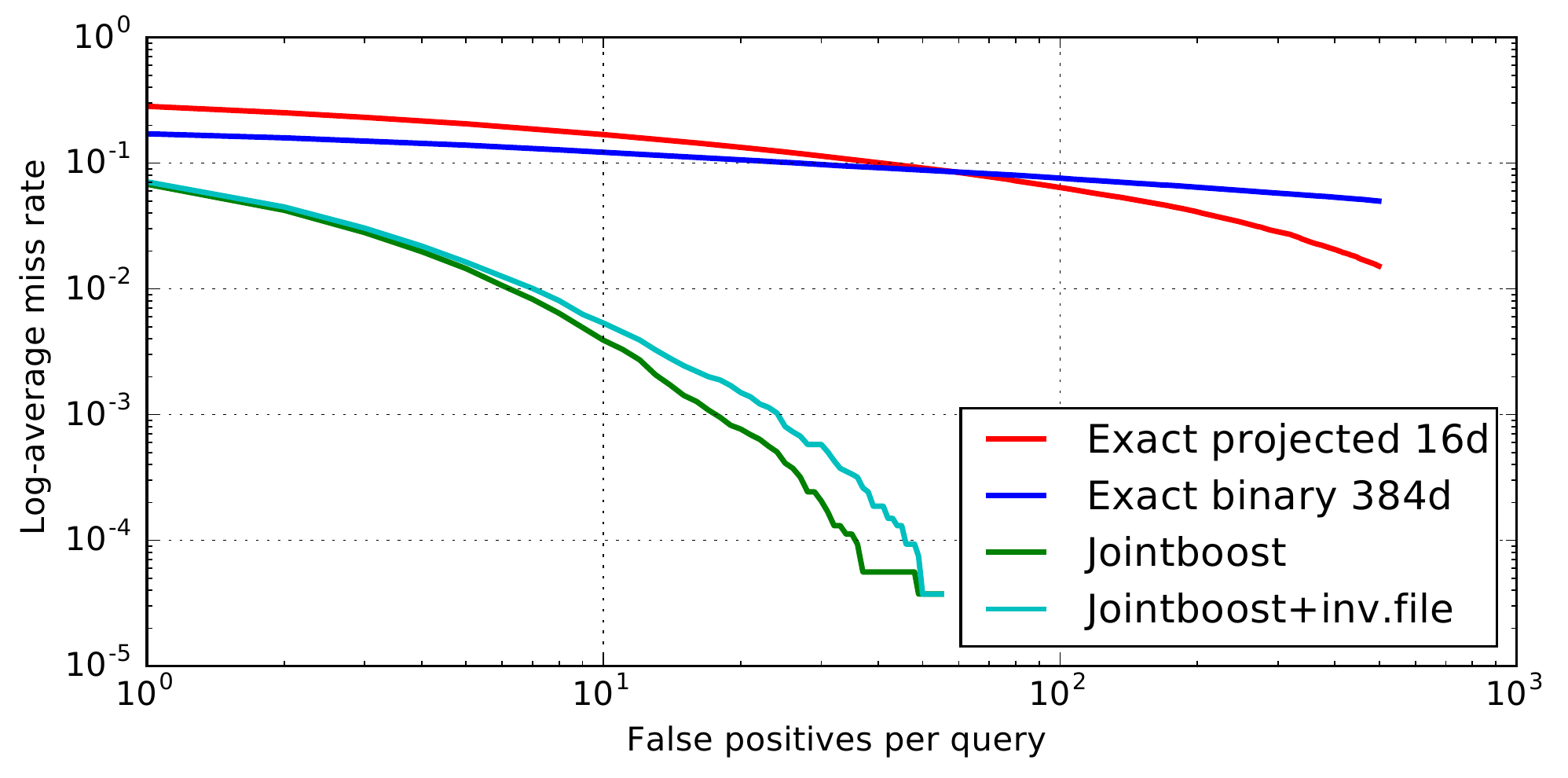}
    \caption{ 
    Average miss rate curve for landmark retrieval of CLA F.
    The proposed jointboost-based approaches deliver significantly reduced average miss rates when compared to the traditional binary and projected descriptors.
    The difference increases with the growing number of retrieved candidates.
    Slightly worse quality of Jointboosting with an inverted file results from quantization errors.
    }
    \label{fig:landmark_retrieval}
\end{figure}

Following the benchmarking methodology of the Caltech Pedestrian Detection Benchmark~\cite{dollar2012pedestrian}, we decided to use the average miss rate and the number of false positives per query to evaluate the retrieval results.
This choice is motivated by the fact that in landmark retrieval only a single landmark class is a true positive.
\reffig{fig:landmark_retrieval} illustrates the results of the proposed and baseline methods.
It shows that significantly fewer candidates need to be retrieved using our approach compared to the baseline to obtain a relevant candidate.
This is confirmed by statistics in \reftab{tab:retrieval}, where both the precision of the first candidate as well as the Mean Reciprocal Rank are superior for LandmarkBoost.

Overall, our results indicate that LandmarkBoost outputs more accurate correspondences, measured both as the reliability of first reported match as well as the position of the true positive within a ranked candidate list.

\begin{table}
\centering
\begin{tabular}{>{\hspace{-3pt}}l >{\hspace{-3pt}}c >{\hspace{-3pt}}c >{\hspace{-3pt}}c >{\hspace{-3pt}}c}
				\toprule
							& Binary 384d & Projected 16d & Jboost & Jboost+inv.file \\ \midrule
				Precision@1 & 0.8081 & 0.6578 & \textbf{0.8592} & 0.8571 \\ \midrule
				MRR & 0.8306 & 0.7163 & \textbf{0.9118} & 0.9096 \\ \bottomrule
\end{tabular}
\caption{Precision@1 and Mean Reciprocal Rank~(MRR) of descriptor retrieval quality.
            The first statistic indicates that the Jointboosting-based methods have a higher probability of returning a correct first candidate.
            The second one shows that the true positive is on average ranked higher by the proposed methods.
            }
\label{tab:retrieval}
\end{table}

\subsection{Pose retrieval}
\label{sec:pose_retrieval}

The aim of this section is to evaluate how the superior landmark retrieval results reported in \refsec{sec:landmark_retrieval} translate into the pose retrieval, the ultimate goal of the localization systems.
2D-3D correspondences retrieved by the matching algorithms are passed to the pose estimation block performing RANSAC on PnP estimates.
The inlier distance threshold and the maximum number of RANSAC iterations (500) are fixed over all evaluations.

The retrieved poses are then evaluated against the ground-truth values refined using a batch visual-inertial least squares optimization as in~\cite{lynen2015get}.
The thresholds of $20$cm and $5$deg are used as a threshold.
All frames of the query datasets are used in this evaluation.

\begin{figure}
    \centering
    \begin{subfigure}[t]{\columnwidth}
      \centering
      \caption{CLA F}
      \vspace{-0.2cm}
      \includegraphics[width=0.85\columnwidth]{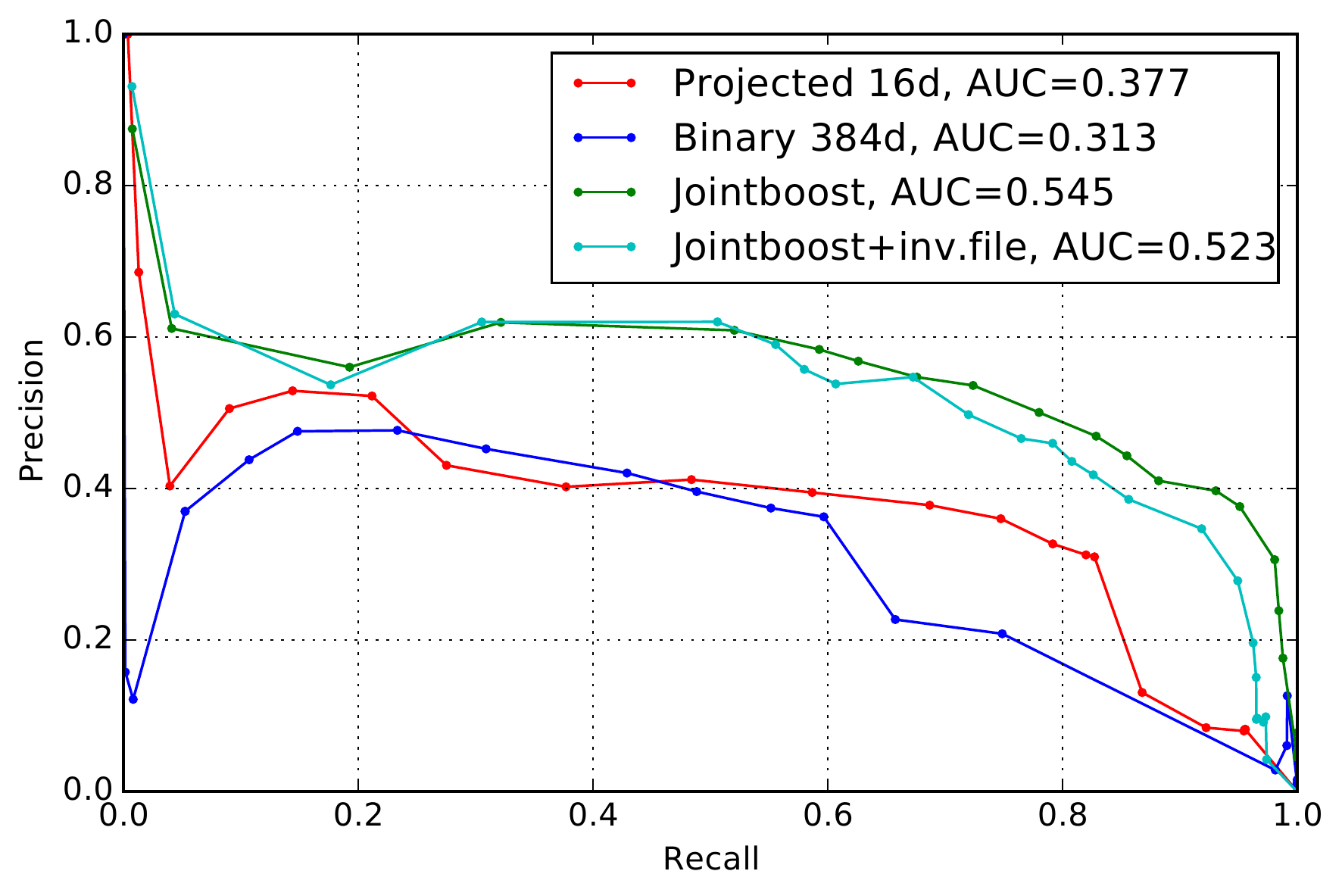}
    \end{subfigure}

    \begin{subfigure}[t]{\columnwidth}
      \centering
      \caption{Zurich Old Town}
      \vspace{-0.2cm}
      \includegraphics[width=0.85\columnwidth]{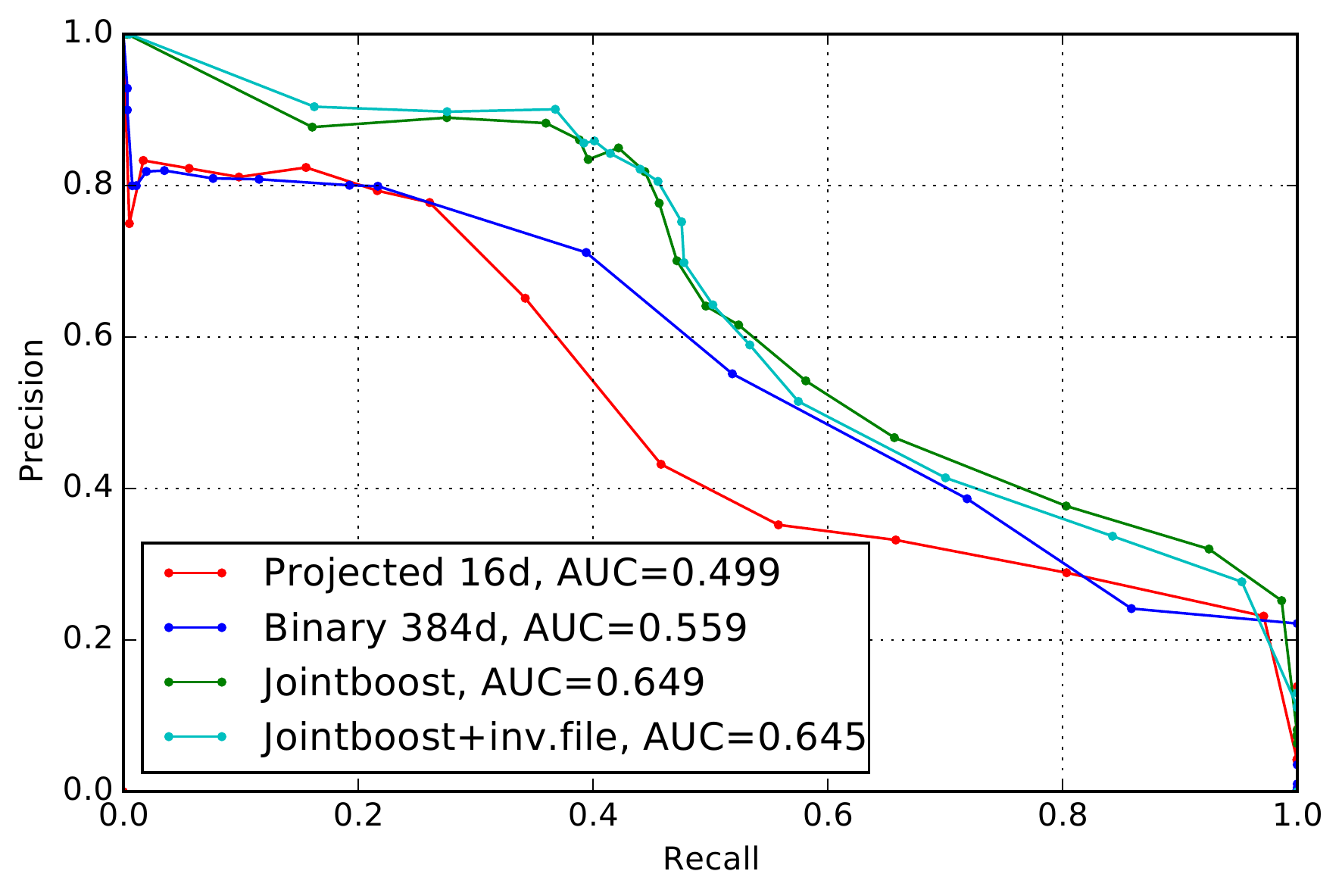}
    \end{subfigure}
    
    \caption{
    The precision-recall curves and the Area Under the Curve (AUC) for 6-DoF pose retrieval.
    Both proposed LandmarkBoost methods outperform the baseline binary and projected descriptor search approaches.
    }
    \label{fig:pr}
\end{figure}

The results are presented in \reffig{fig:pr}.
LandmarkBoost methods bring a boost in performance, especially for the high recall part of the curve, where the precision of the conventional methods drops rapidly.
This permits to correctly localize more frames, without sacrificing precision of the output.

\subsection{Matching runtime evaluation}
\label{sec:runtime}

\begin{figure}[h!]
    \centering
    \includegraphics[width=1.0\columnwidth]{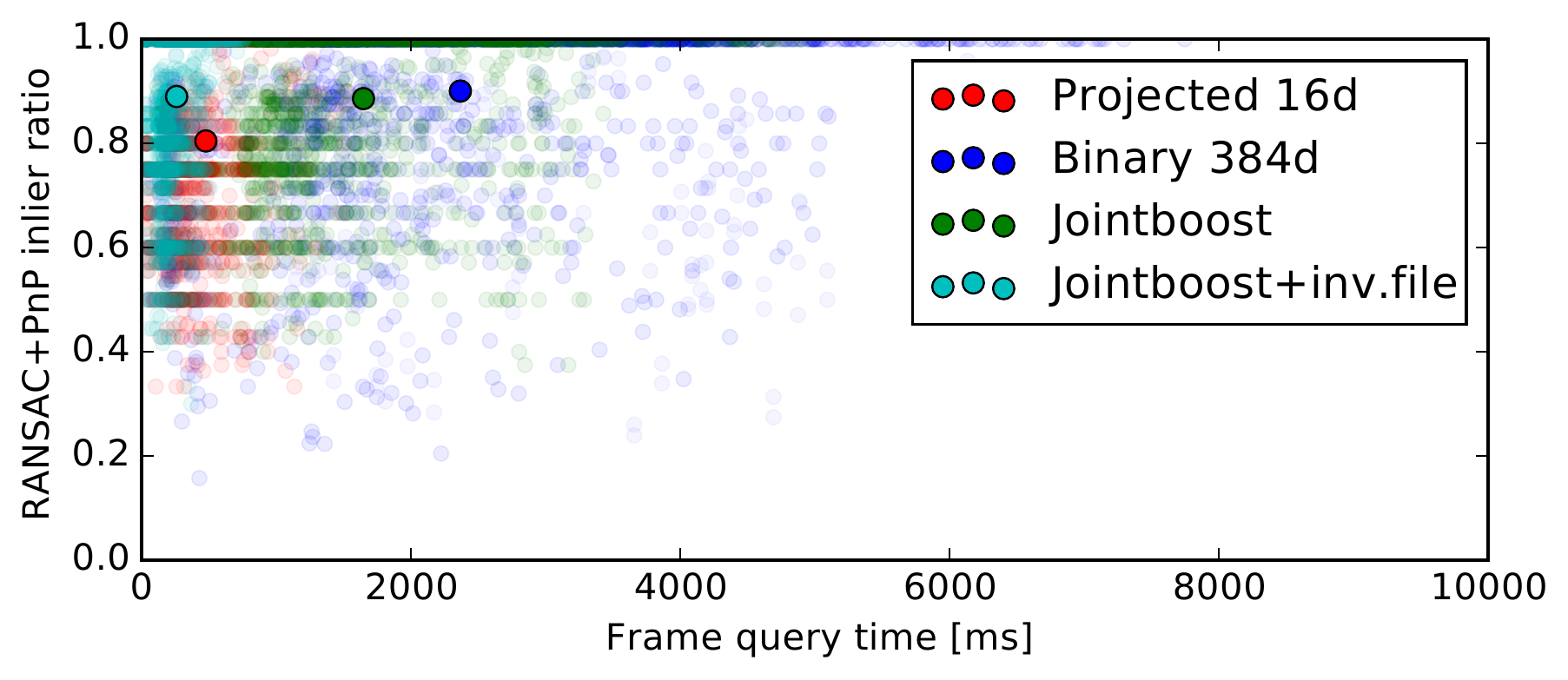}
    \caption{
    Runtime vs average inlier ratio.
    Mean runtime-inlier ratio pairs for each method are marked with full intensity markers.
    The figure presents the trade-off between the quality of landmark matches and the query time (single core of Intel i7-7820X@3.6GHz).
    The proposed Jointboost method delivers inlier ratios close to the baseline, but with about 30\% shorter runtime.
    Using an inverted file for classification (see \refsec{sec:classification}) further reduces the computational cost of LandmarkBoost.
    }
    \label{fig:runtime_inlier}
    \vspace{-7mm}
\end{figure}

In this section, we present an evaluation of the classification (or search) runtime.
We decided to assess the runtime with respect to the inlier ratio to illustrate the trade-off between the matching quality and computational requirements.
As shown in \reffig{fig:runtime_inlier}, the proposed methods deliver the quality that is on par with the binary descriptor retrieval, but at a reduced query time, particularly for the classification using a set of candidates from the quantizer inverted file.

We believe that the runtime of Jointboost classification can be further reduced if necessary. 
The classifier evaluation consists of a series of binary tests and simple score arithmetic.
These operations can be easily parallelized, further optimized using CPU vector instructions or even deployed on a GPU.

\section{Conclusions}

In this paper, we have presented LandmarkBoost, a novel approach for precise, metric localization.
We have shown it is possible to construct a boosted classifier that operates on a large number of classes and delivers better performance than conventional, search based methods.
The classifier incorporates not only the local point features, but also the visual context by mining for useful information that consistently reappears with a given landmark.
This way, we achieve a higher level of robustness against visual aliasing or appearance changes compared to raw binary features.
Finally, we have presented a comprehensive evaluation that demonstrates superior descriptor and pose retrieval quality and shows real-time capability for robotic applications.

While this paper focuses on binary descriptors and visual context based on neighboring keypoints, LandmarkBoost can easily accommodate different data types while remaining very efficient.
The proposed formulation is ready to work with other feature descriptors or region descriptors.
We also believe that extending the framework to other kinds of data, such as semantics or object labels, might further improve the performance when compared to conventional descriptor search-based localization.




\bibliographystyle{IEEEtran} 
\bibliography{references}

\begin{thebibliography}{10}
\providecommand{\url}[1]{#1}
\csname url@rmstyle\endcsname
\providecommand{\newblock}{\relax}
\providecommand{\bibinfo}[2]{#2}
\providecommand\BIBentrySTDinterwordspacing{\spaceskip=0pt\relax}
\providecommand\BIBentryALTinterwordstretchfactor{4}
\providecommand\BIBentryALTinterwordspacing{\spaceskip=\fontdimen2\font plus
\BIBentryALTinterwordstretchfactor\fontdimen3\font minus
  \fontdimen4\font\relax}
\providecommand\BIBforeignlanguage[2]{{%
\expandafter\ifx\csname l@#1\endcsname\relax
\typeout{** WARNING: IEEEtran.bst: No hyphenation pattern has been}%
\typeout{** loaded for the language `#1'. Using the pattern for}%
\typeout{** the default language instead.}%
\else
\language=\csname l@#1\endcsname
\fi
#2}}

\bibitem{leutenegger2011brisk}
S.~Leutenegger, M.~Chli, and R.~Y. Siegwart, ``Brisk: Binary robust invariant
  scalable keypoints,'' in \emph{ICCV}, 2011.

\bibitem{alahi2012freak}
A.~Alahi, R.~Ortiz, and P.~Vandergheynst, ``Freak: Fast retina keypoint,'' in
  \emph{Computer vision and pattern recognition (CVPR), 2012 IEEE conference
  on}, 2012.

\bibitem{cummins2008fab}
M.~Cummins and P.~Newman, ``Fab-map: Probabilistic localization and mapping in
  the space of appearance,'' \emph{The International Journal of Robotics
  Research}, vol.~27, no.~6, 2008.

\bibitem{norouzi2012fast}
M.~Norouzi, A.~Punjani, and D.~J. Fleet, ``Fast search in hamming space with
  multi-index hashing,'' in \emph{Computer Vision and Pattern Recognition
  (CVPR), 2012 IEEE Conference on}, 2012.

\bibitem{lynen2014placeless}
S.~Lynen, M.~Bosse, P.~Furgale, and R.~Siegwart, ``Placeless
  place-recognition,'' in \emph{3DV}, 2014.

\bibitem{sattler2011fast}
T.~Sattler, B.~Leibe, and L.~Kobbelt, ``Fast image-based localization using
  direct 2d-to-3d matching,'' in \emph{Computer Vision (ICCV), 2011 IEEE
  International Conference on}, 2011.

\bibitem{middelberg2014scalable}
S.~Middelberg, T.~Sattler, O.~Untzelmann, and L.~Kobbelt, ``Scalable 6-dof
  localization on mobile devices,'' in \emph{European conference on computer
  vision}, 2014.

\bibitem{lynen2015get}
S.~Lynen, T.~Sattler, M.~Bosse, J.~A. Hesch, M.~Pollefeys, and R.~Siegwart,
  ``Get out of my lab: Large-scale, real-time visual-inertial localization.''
  in \emph{RSS}, 2015.

\bibitem{fankhauser2016collaborative}
P.~Fankhauser, M.~Bloesch, P.~Kr{\"u}si, R.~Diethelm, M.~Wermelinger,
  T.~Schneider, M.~Dymczyk, M.~Hutter, and R.~Siegwart, ``Collaborative
  navigation for flying and walking robots,'' in \emph{Intelligent Robots and
  Systems (IROS), 2016 IEEE/RSJ International Conference on}, 2016.

\bibitem{jegou2008hamming}
H.~Jegou, M.~Douze, and C.~Schmid, ``Hamming embedding and weak geometric
  consistency for large scale image search,'' in \emph{European conference on
  computer vision}, 2008.

\bibitem{zeisl2015camera}
B.~Zeisl, T.~Sattler, and M.~Pollefeys, ``Camera pose voting for large-scale
  image-based localization,'' in \emph{Computer Vision (ICCV), 2015 IEEE
  International Conference on}, 2015.

\bibitem{gehrig2017visual}
M.~Gehrig, E.~Stumm, T.~Hinzmann, and R.~Siegwart, ``Visual place recognition
  with probabilistic voting,'' in \emph{ICRA}, 2017.

\bibitem{csurka2004visual}
G.~Csurka, C.~Dance, L.~Fan, J.~Willamowski, and C.~Bray, ``Visual
  categorization with bags of keypoints,'' in \emph{Workshop on statistical
  learning in computer vision, ECCV}, vol.~1, no. 1-22, 2004.

\bibitem{jegou2010aggregating}
H.~J{\'e}gou, M.~Douze, C.~Schmid, and P.~P{\'e}rez, ``Aggregating local
  descriptors into a compact image representation,'' in \emph{Computer Vision
  and Pattern Recognition (CVPR), 2010 IEEE Conference on}, 2010.

\bibitem{irschara2009structure}
A.~Irschara, C.~Zach, J.-M. Frahm, and H.~Bischof, ``From structure-from-motion
  point clouds to fast location recognition,'' in \emph{Computer Vision and
  Pattern Recognition, 2009. CVPR 2009. IEEE Conference on}, 2009.

\bibitem{oliva2001modeling}
A.~Oliva and A.~Torralba, ``Modeling the shape of the scene: A holistic
  representation of the spatial envelope,'' \emph{International journal of
  computer vision}, 2001.

\bibitem{gordo2016deep}
A.~Gordo, J.~Almaz{\'a}n, J.~Revaud, and D.~Larlus, ``Deep image retrieval:
  Learning global representations for image search,'' in \emph{European
  Conference on Computer Vision}, 2016.

\bibitem{weyand2016planet}
T.~Weyand, I.~Kostrikov, and J.~Philbin, ``Planet-photo geolocation with
  convolutional neural networks,'' in \emph{European Conference on Computer
  Vision}, 2016.

\bibitem{milford2012seqslam}
M.~J. Milford and G.~F. Wyeth, ``Seqslam: Visual route-based navigation for
  sunny summer days and stormy winter nights,'' in \emph{Robotics and
  Automation (ICRA), 2012 IEEE International Conference on}, 2012.

\bibitem{clark2017vidloc}
R.~Clark, S.~Wang, A.~Markham, N.~Trigoni, and H.~Wen, ``Vidloc: A deep
  spatio-temporal model for 6-dof video-clip relocalization,'' in
  \emph{Proceedings of IEEE Conference on Computer Vision and Pattern
  Recognition (CVPR)}, 2017.

\bibitem{sunderhauf2013we}
N.~S{\"u}nderhauf, P.~Neubert, and P.~Protzel, ``Are we there yet? challenging
  seqslam on a 3000 km journey across all four seasons,'' in \emph{Proc. of
  Workshop on Long-Term Autonomy, IEEE International Conference on Robotics and
  Automation (ICRA)}, 2013.

\bibitem{mei2010closing}
C.~Mei, G.~Sibley, and P.~Newman, ``Closing loops without places,'' in
  \emph{Intelligent Robots and Systems (IROS), 2010 IEEE/RSJ International
  Conference on}, 2010.

\bibitem{stumm2016building}
E.~S. Stumm, C.~Mei, and S.~Lacroix, ``Building location models for visual
  place recognition,'' \emph{The International Journal of Robotics Research},
  vol.~35, no.~4, 2016.

\bibitem{stumm2016robust}
E.~Stumm, C.~Mei, S.~Lacroix, J.~Nieto, M.~Hutter, and R.~Siegwart, ``Robust
  visual place recognition with graph kernels,'' in \emph{Proceedings of the
  IEEE Conference on Computer Vision and Pattern Recognition}, 2016.

\bibitem{lowry2018logos}
S.~Lowry and H.~Andreasson, ``Logos: Local geometric support for high-outlier
  spatial verification,'' in \emph{2018 IEEE International Conference on
  Robotics and Automation (ICRA)}, May 2018.

\bibitem{loquercio2017efficient}
A.~Loquercio, M.~Dymczyk, B.~Zeisl, S.~Lynen, I.~Gilitschenski, and
  R.~Siegwart, ``Efficient descriptor learning for large scale localization,''
  in \emph{Robotics and Automation (ICRA), 2017 IEEE International Conference
  on}, 2017.

\bibitem{zhang2011image}
Y.~Zhang, Z.~Jia, and T.~Chen, ``Image retrieval with geometry-preserving
  visual phrases,'' in \emph{Computer Vision and Pattern Recognition (CVPR),
  2011 IEEE Conference on}, 2011.

\bibitem{yuan2007discovery}
J.~Yuan, Y.~Wu, and M.~Yang, ``Discovery of collocation patterns: from visual
  words to visual phrases,'' in \emph{Computer Vision and Pattern Recognition,
  2007. CVPR'07. IEEE Conference on}, 2007.

\bibitem{mcmanus2014scene}
C.~McManus, B.~Upcroft, and P.~Newmann, ``Scene signatures: Localised and
  point-less features for localisation,'' in \emph{Proceedings of Robotics
  Science and Systems (RSS)}, 2014.

\bibitem{linegar2016made}
C.~Linegar, W.~Churchill, and P.~Newman, ``Made to measure: Bespoke landmarks
  for 24-hour, all-weather localisation with a camera,'' in \emph{Robotics and
  Automation (ICRA), 2016 IEEE International Conference on}, 2016.

\bibitem{torralba2004sharing}
A.~Torralba, K.~P. Murphy, and W.~T. Freeman, ``Sharing features: efficient
  boosting procedures for multiclass object detection,'' in \emph{Computer
  Vision and Pattern Recognition, 2004. CVPR 2004. Proceedings of the 2004 IEEE
  Computer Society Conference on}, 2004.

\bibitem{friedman2000additive}
J.~Friedman, T.~Hastie, R.~Tibshirani, \emph{et~al.}, ``Additive logistic
  regression: a statistical view of boosting (with discussion and a rejoinder
  by the authors),'' \emph{The annals of statistics}, vol.~28, no.~2, 2000.

\bibitem{philbin2010descriptor}
J.~Philbin, M.~Isard, J.~Sivic, and A.~Zisserman, ``Descriptor learning for
  efficient retrieval,'' in \emph{European Conference on Computer Vision},
  2010.

\bibitem{sung1998example}
K.-K. Sung and T.~Poggio, ``Example-based learning for view-based human face
  detection,'' \emph{IEEE Transactions on pattern analysis and machine
  intelligence}, vol.~20, no.~1, 1998.

\bibitem{torralba2007sharing}
A.~Torralba, K.~P. Murphy, and W.~T. Freeman, ``Sharing visual features for
  multiclass and multiview object detection,'' \emph{IEEE Transactions on
  Pattern Analysis and Machine Intelligence}, vol.~29, no.~5, 2007.

\bibitem{shotton2006textonboost}
J.~Shotton, J.~Winn, C.~Rother, and A.~Criminisi, ``Textonboost: Joint
  appearance, shape and context modeling for multi-class object recognition and
  segmentation,'' in \emph{European conference on computer vision}, 2006.

\bibitem{kneip2014opengv}
L.~Kneip and P.~Furgale, ``Opengv: A unified and generalized approach to
  real-time calibrated geometric vision,'' in \emph{Robotics and Automation
  (ICRA), 2014 IEEE International Conference on}, 2014.

\bibitem{krahenbuhl2011efficient}
P.~Kr{\"a}henb{\"u}hl and V.~Koltun, ``Efficient inference in fully connected
  crfs with gaussian edge potentials,'' in \emph{Advances in neural information
  processing systems}, 2011.

\bibitem{schneider2018maplab}
T.~Schneider, M.~Dymczyk, M.~Fehr, K.~Egger, S.~Lynen, I.~Gilitschenski, and
  R.~Siegwart, ``maplab: An open framework for research in visual-inertial
  mapping and localization,'' \emph{ICRA}, 2018.

\bibitem{opencv_library}
G.~Bradski, ``{The OpenCV Library},'' \emph{Dr. Dobb's Journal of Software
  Tools}, 2000.

\bibitem{elseberg2012comparison}
J.~Elseberg, S.~Magnenat, R.~Siegwart, and A.~N{\"u}chter, ``Comparison of
  nearest-neighbor-search strategies and implementations for efficient shape
  registration,'' \emph{Journal of Software Engineering for Robotics}, 2012.

\bibitem{dymczyk2015gist}
M.~Dymczyk, S.~Lynen, T.~Cieslewski, M.~Bosse, R.~Siegwart, and P.~Furgale,
  ``The gist of maps-summarizing experience for lifelong localization,'' in
  \emph{ICRA}, 2015.

\bibitem{dymczyk2015keep}
M.~Dymczyk, S.~Lynen, M.~Bosse, and R.~Siegwart, ``Keep it brief: Scalable
  creation of compressed localization maps,'' in \emph{IROS}, 2015.

\bibitem{dollar2012pedestrian}
P.~Dollar, C.~Wojek, B.~Schiele, and P.~Perona, ``Pedestrian detection: An
  evaluation of the state of the art,'' \emph{IEEE transactions on pattern
  analysis and machine intelligence}, vol.~34, no.~4, 2012.

\end{thebibliography}

\end{document}